\documentclass[]{assets/template}
\usepackage[utf8]{inputenc}
\usepackage{setspace}
\usepackage{xcolor}
\usepackage{color}
\usepackage{amssymb, amsmath, amsthm, mathrsfs, mathtools, bbm, dsfont}

\usepackage{graphicx}
\usepackage{caption}
\usepackage{subcaption}
\usepackage{booktabs}
\usepackage{threeparttable}
\usepackage{multirow}
\usepackage{makecell}
\usepackage{tabularx}
\usepackage{colortbl}
\usepackage{arydshln}
\usepackage{float}
\usepackage{wrapfig}

\usepackage{algorithm}
\usepackage{algpseudocode}

\usepackage{enumitem}
\usepackage{tcolorbox}
\usepackage{markdown}
\usepackage{pifont}
\usepackage{fontawesome5}
\usepackage{listings}
\usepackage{makecell}

\usepackage{tikz}
\usetikzlibrary{fit, calc}

\usepackage{natbib}
\usepackage{hyperref}
\usepackage{url}
\usepackage{cleveref}

\usepackage{bxcoloremoji}

\addtocontents{toc}{\protect\setcounter{tocdepth}{0}}

\definecolor{metagreen}{HTML}{2E8B57} 

\newcommand{\M}{Mnemis}
\newcolumntype{Y}{>{\centering\arraybackslash}X}

\title{\M: Dual-Route Retrieval on Hierarchical Graphs for Long-Term LLM Memory}

\author{
Zihao Tang, Xin Yu*, Ziyu Xiao, Zengxuan Wen, Zelin Li,
Jiaxi Zhou, Hualei Wang, Haohua Wang,
Haizhen Huang, Denvy Deng, Feng Sun, Qi Zhang \\
  Microsoft\\
  $^\star$\texttt{Corresponding Author: xinyu2@microsoft.com}\\
}

\runningtitle{\M: Dual-Route Retrieval on Hierarchical Graphs for Long-Term LLM Memory}

\begin{document}
\begin{abstract}

AI Memory, specifically how models organizes and retrieves historical messages,  becomes increasingly valuable to Large Language Models (LLMs), yet existing methods (RAG and Graph-RAG) primarily retrieve memory through similarity-based mechanisms. While efficient, such System-1-style retrieval struggles with scenarios that require global reasoning or comprehensive coverage of all relevant information. In this work, We propose \M, a novel memory framework that integrates System-1 similarity search with a complementary System-2 mechanism, termed Global Selection. \M~organizes memory into a base graph for similarity retrieval and a hierarchical graph that enables top-down, deliberate traversal over semantic hierarchies. By combining the complementary strength from both retrieval routes, \M~retrieves memory items that are both semantically and structurally relevant. \M~achieves state-of-the-art performance across all compared methods on long-term memory benchmarks, scoring 93.9 on LoCoMo and 91.6 on LongMemEval-S using GPT-4.1-mini. \\

\faGlobe~ \textbf{Project}: \href{https://github.com/microsoft/Mnemis}{https://github.com/microsoft/Mnemis} \\
\faCalendar~ \textbf{Date}: 6 Jan 2026

\end{abstract}

\maketitle
\section{Introduction}
With the rapid advancement of Large Language Models (LLMs), there is a growing trend to integrate memory mechanisms to support long-term interactions \citep{lewis2020retrieval,ouyang2025reasoningbank,behrouz2024titans,DBLP:conf/acl/WuCLLZLKW25,DBLP:journals/corr/abs-2508-11184,tang2024modelgptunleashingllmscapabilities,liang2025pass1selfplayvariationalproblem}. As LLMs shift from text generators to persistent interactive agents, the ability to organize and retrieve past interactions becomes increasingly valuable. The prevailing research paradigm is based on retrieval-augmented generation (RAG). Inspired by human episodic memory \citep{tulving1972episodic}, these methods (\textit{e.g.} SeCom \citep{pan2025secom}, Memory-R1 \citep{yan2025memory}) explicitly store historical messages (\textit{i.e., Episodes}) and retrieve only the most relevant pieces \citep{arslan2024survey,lewis2020retrieval}. This design alleviates the computational and latency issues of long-context models and keeps the input compact and focused. However, its effectiveness critically depends on retrieval quality.

Recent work on graph-based RAG (Graph-RAG) extends RAG by incorporating concepts from semantic memory \citep{tulving1972episodic}. Graph-RAG extracts memory segments, \textit{e.g.} \textit{Entities} (key figures, objects, or concepts) and \textit{Edges} (events or relationships connecting Entities) and organizes memory into a structured graph, as exemplified by methods such as GraphRAG \citep{edge2024local}, Nemori \citep{nan2025nemoriselforganizingagentmemory}, Mem0 \citep{chhikara2025mem0buildingproductionreadyai},  and Zep \citep{rasmussen2025zeptemporalknowledgegraph}. These methods highlight essential information and enable more effective and semantically meaningful retrieval.

\begin{figure}[!th]
    \centering
    \vspace{-0.5cm}
    \includegraphics[width=\linewidth]{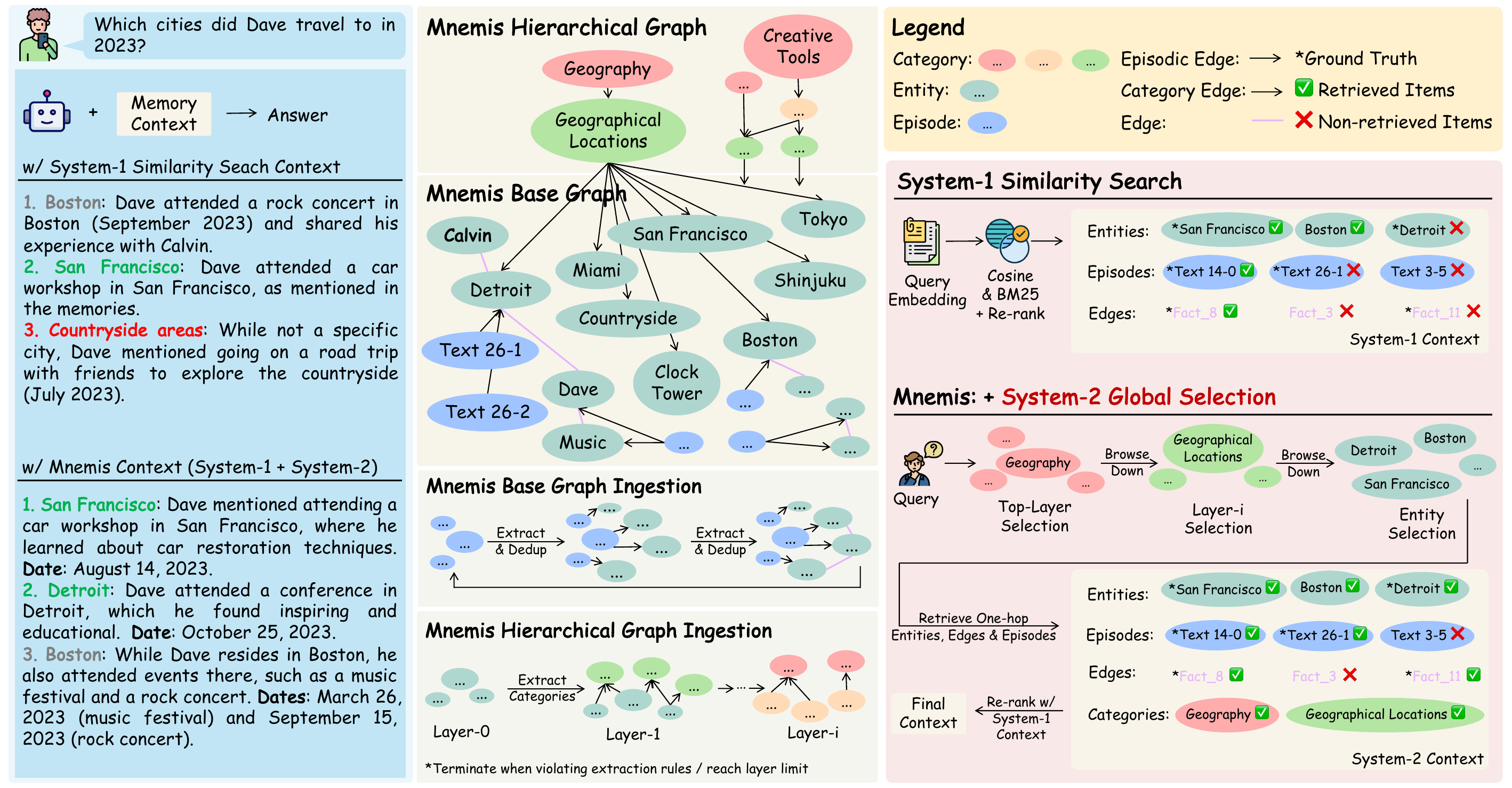}
    \caption{Framework of \M~together with the workflow of base graph ingestion, hierarchical graph ingestion and search. Left is a real case from LoCoMo.}
    \label{fig:fig_intro}
\end{figure}

Although Graph-RAG-based methods mark an important step toward structured memory, their retrieval remains largely similarity-driven, selecting Episodes, Entities, or Edges via text matching (BM25) or embedding similarity (cosine). This approach is fast and effective, and resembles the System-1 process in dual-process theory \citep{kahneman2011thinking}, but becomes limited when queries require global reasoning or comprehensive coverage of all relevant information. Although recent research has explored iterative generation of sub-queries to mitigate this issue, such methods still fall short for questions that require a broader perspective \citep{wang2025chain-of-retrieval,jin2025searchr1trainingllmsreason}. For example, consider the query \textit{"Which cities did Dave travel to in 2023?"} from LoCoMo Benchmark \citep{maharana2024evaluating}, as shown in \cref{fig:fig_intro}. The mention \textit{"attended a conference in Detroit."} is buried in a long message and has only a weak semantic relation to the user query. Moreover, generating effective sub-queries is challenging as the model lacks a global view of the memory to determine how the original query should be meaningfully expanded.

Recalling how human approach such questions, it can be naturally addressed using a semantic hierarchy. We can begin with a high-level concept (\textit{e.g.} city), enumerate all the cities we have visited and verify them one-by-one. This kind of solution operates over a global view of memory and naturally avoids the need for sub-query generation, reflecting a structured process characteristic of System-2 reasoning \citep{kahneman2011thinking}.

Inspired by this observation, we propose an analogous mechanism, called \textit{Global Selection}, which constructs a hierarchical graph that provides a complete, global, and structured view of the entire memory, mimicking human semantic hierarchies. It allows models to perform top-down, deliberate memory scanning within it. In this example, Global Selection can start from the top layer and follow the path "Geography" → "Geographical Locations" → "Detroit" to retrieve the relevant information.

In practice, real-world queries often benefit from combining both the System-1 and System-2 processes, as they operate through different retrieval patterns. Motivated by this, we present \M, a novel and effective framework to organize and retrieve AI memory. \M~comprises two storage components: a base graph and a hierarchical graph, and two corresponding retrieval routes: System-1 similarity search and System-2 global selection. The base graph, similar to prior Graph-RAG designs, extracts Entities and Edges from history texts (Episodes) to support similarity-based retrieval. We refine the extraction pipeline to increase extraction fields and improve extraction quality.
In contrast, the hierarchical graph prompts LLMs to categorize Entities into higher-level Categories through bottom-up. This process follows three key principles: (1) Minimum Concept Abstraction: each Category should faithfully capture the shared features of its child nodes. It should be specific enough to be informative, yet sufficiently general to support abstraction; (2) Many-to-Many Mapping: one child node can be assigned to multiple Categories to represent its different semantic facets; and (3) Compression Efficiency Constraint: one Category must contain at least $n$ children and higher layers must contain no more Categories than lower layers (applied from layer 2 onward).

When a query arrives, the similarity search route conducts a semantic search based on embeddings and text similarity, while the global search performs a top-down selection through the hierarchical graph, layer by layer. Down to the lowest level, the LLM first selects all relevant entities and then retrieves all edges, entities and episodes connected to them.
These two routes capture complementary signals: System-1 provides fine-grained semantic similarity evidence, while System-2 retrieves structurally relevant items that may be semantically distant yet relationally important.
By combining and re-ranking the union of both routes, \M~achieves SOTA performance across all compared methods on long-term memory benchmarks, scoring 93.9 on LoCoMo and 91.6 on LongMemEval-S using GPT-4.1-mini.
Our contributions can be summarized as below:
\begin{itemize}
    \item We introduce \M, a novel framework that integrates System-1 similarity search with System-2 global selection to perform both semantic retrieval and deliberate, top-down reasoning over memory;
    \item We improve the base graph extraction and construct a hierarchical graph for global selection, guided by Minimum Concept Abstraction, Many-to-Many Mapping, and Compression Efficiency Constraint to maintain hierarchical quality;
    \item We perform comprehensive experiments to demonstrate the effectiveness of \M. \M~achieves SOTA performance across all compared methods on long-term memory benchmarks, scoring 93.9 on LoCoMo and 91.6 on LongMemEval-S using GPT-4.1-mini.
\end{itemize}
\section{\M~Methodology}
To achieve effective memory organization, \M~constructs two major components: a base graph and a hierarchical graph and two key memory retrieval mechanisms: System-1 Similarity Search and System-2 Global Selection. We implement \M~based on Graphiti\footnote{\url{https://github.com/getzep/graphiti}}.

\subsection{Base Graph}

The base graph stores historical messages and captures detailed information, enabling the model to perform System-1 Similarity Search, \textit{i.e.}, retrieving semantically relevant histories. It consists of four components: Episodes, Entities, Edges and Episodic Edges.

\textbf{Episodes.} Each episode is a piece of raw historical text. It is encoded into an \texttt{episode\_embedding} for similarity-based retrieval. Its timestamp is recorded at \texttt{valid\_at}.

\textbf{Entities.} An entity is any concrete person, organization, place, object, event, or well-defined concept. Each entity includes \texttt{name}, \texttt{summary}, \texttt{tag}, and \texttt{episode\_idx}. The \texttt{summary} provides a concise contextual description, the \texttt{tag} specifies its type or role, and \texttt{episode\_idx} tracks the episodes it appears. We encode \texttt{name} and \texttt{summary} into corresponding embeddings for flexible search.

\textbf{Edges.} An edge is a verifiable statement describing a meaningful relationship, action, or state involving one or more specified entities within a defined temporal or contextual scope. Each edge connects two entities through a \texttt{fact}, which is encoded as a \texttt{fact\_embedding}. Additionally, \texttt{valid\_at} and \texttt{invalid\_at} specify the time span during which the edge is considered valid.

\textbf{Episodic Edges.} An episodic edge links entities to all episodes where they appear. It is utilized during global search to retrieve all episodes associated with selected entities.

The ingestion of the base graph is conducted incrementally: new inputs are first formatted into Episodes. Based on their timestamps, recent Episodes will be retrieved to provide additional context. During extraction, the LLM first identifies entity names from both the current and recent Episodes, followed by a reflection process to capture omitted entity names. These names are then de-duplicated against existing entities in memory, using a combination of full-text search and similarity search over the \texttt{name\_embedding}. After de-duplication, each entity’s \texttt{summary}, \texttt{tag}, and \texttt{episode\_idx} are extracted according to the episode context. Subsequently, Edges are extracted using both Episodes and Entities as contextual inputs, followed by reflection and de-duplication steps analogous to those used in entity extraction.
To ensure the robustness of base graph ingestion, we forcibly rule-extract the speaker as an entity.

\subsection{Hierarchical Graph}

\begin{wrapfigure}{r}{0.5\linewidth}
    \centering
    \includegraphics[width=\linewidth]{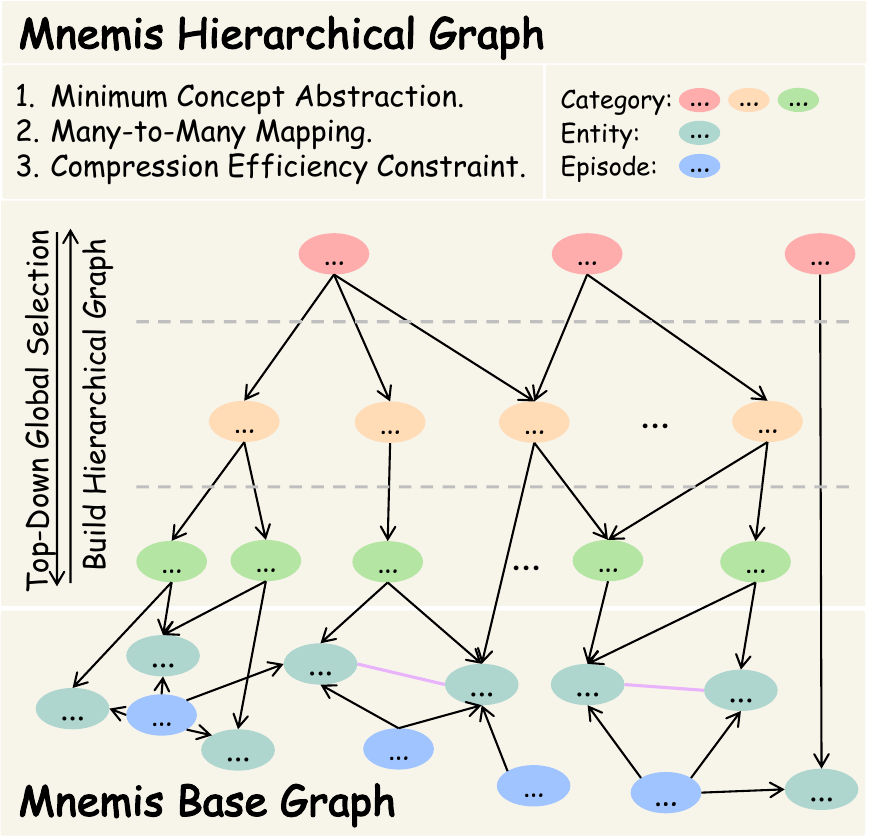}
    \vspace{-.7cm}
    \caption{\M~Hierarchical Graph Overview.}
    \vspace{-1.3cm}
    \label{fig:fig_hierarchy}
\end{wrapfigure}

The hierarchical graph abstracts Entities (layer 0) into multi-level Categories, enabling the LLM to perform System-2 Global Selection. The structure consists of two components, as shown in \cref{fig:fig_hierarchy}.

\vspace{-.2cm}
\vspace{-.3cm}
\par\noindent

\textbf{Category Nodes (Categories).} A category represents an abstract, high-level concept derived from lower-layer categories (or entities at layer 0). It shares the same core fields as an Entity, with an additional attribute \texttt{layer} indicating its position within the hierarchical graph.

\textbf{Category Edges.} A category edge links a higher-layer category to its child nodes (either lower-layer categories or entities). These edges define the hierarchical organization of the graph and support the top-down traversal process in global selection.

The ingestion of hierarchical graph is governed by three key design principles:

\textbf{Minimum Concept Abstraction.} While categories are intended to capture the shared semantics of their child nodes, we explicitly prompt the LLM to perform \textit{minimal abstraction}. The resulting category should remain sufficiently specific to preserve informative detail, leaving room for broader generalizations at higher layers.

\textbf{Many-to-Many Mapping.} Unlike conventional tree-structured hierarchies, \M~permits lower-layer nodes to belong to multiple higher-layer categories. This design allows the hierarchy to represent different semantic facets of each node, enabling retrieval from multiple perspectives depending on the query.

\textbf{Compression Efficiency Constraint.} To ensure the efficiency of System-2 Global Selection, the hierarchy is regulated by two complementary mechanisms: (1) the \textit{compression ratio} $n$ and (2) the \textit{node count reduction rule}, which takes effect from layer~2 onward.

The compression ratio constrains the hierarchy at the category level. Each category must contain at least $n$ child nodes. An exception is made for nodes that cannot be naturally merged with others; such nodes are directly promoted to the next layer as standalone categories, encouraging meaningful aggregation while preventing overly fine-grained or trivial categories.

The node count reduction rule, in contrast, constrains the hierarchy at the layer level: each upper layer must contain no more nodes than the layer beneath it, ensuring progressive abstraction across layers. If this rule is violated, \textit{e.g.}, when multiple nodes are promoted directly without merging and the result layer is oversized, the ingestion process is terminated to maintain hierarchical balance.

Guided by the principles above, the hierarchical graph is constructed layer by layer. At layer $i$, all nodes from layer $i-1$ are first retrieved. Category names are then generated, and lower-layer nodes are assigned to these categories using their names and tags as contextual information. The construction process terminates when either the compression efficiency constraints are violated or the maximum layer limit is reached.

\textbf{Practical Scalability.} During base graph ingestion, we include a fallback design that explicitly extracts speakers as dedicated entities. Accordingly, We aggregate them into a layer-1 Speaker category in the hierarchical graph as shown in \cref{sec:prompt}. Speaker is further organized into higher-level categories, introducing additional retrieval paths and thereby improving recall and reducing sensitivity to local categorization errors. For example, queries such as user’s allergy can be resolved either through health-related categories or through the speaker branch. When the base graph is updated, the hierarchical graph should be updated accordingly. Currently, we periodically rebuild the hierarchical graph for simplicity and leave optimization for future work.

\subsection{Memory Retrieval Mechanisms}
Basically, \M~contains two major memory retrieval routes: System-1 Similarity Search and System-2 Global Selection. Given a user query, \M~retrieves Episodes, Entities and Edges, formatting them into a context and prompts LLM to get the final answer.

\textbf{System-1 Similarity Search.} This route retrieves the top-$k$ Episodes, Entities, and Edges, providing fast and effective retrieval based on semantic similarity. It operates through two complementary methods: embedding search, which retrieves relevant items by computing cosine similarity between the query embedding and the corresponding embeddings (\texttt{summary\_embedding} for Entities, \texttt{fact\_embedding} for Edges, and \texttt{episode\_embedding} for Episodes); and full-text search, which retrieves relevant components using BM25 over textual content (\texttt{content} for Episodes, \texttt{name} and \texttt{summary} for Entities, and \texttt{fact} for Edges). These two results are then merged and re-ranked using reciprocal rank fusion (RRF) \citep{cormack2009reciprocal}, which computes a fusion score by summing the reciprocals of each candidate’s ranks and orders candidates in descending score. Episodes, Entities, and Edges are re-ranked separately. A higher $\text{RRFScore}(x)$ corresponds to a higher rank for the candidate. The re-ranked results are then truncated to the top-$k$ items according to the predefined search budget.

\textbf{System-2 Global Selection.} This route enables deliberate, top-down exploration of memory through the hierarchical graph. Because the process is primarily structure-driven and the selection at each layer is fully determined by the LLM, no strict top-$k$ constraint is applied. Starting from the top layer, the LLM uses category names and tags to select relevant Categories based on the user query and progressively browses down the hierarchy layer by layer. At the lowest level, all relevant entities are first selected. \M~then retrieves all episodes and edges directly connected to these entities, along with the entities linked through those edges.

\textbf{Re-ranking.} After executing both retrieval routes, we apply a re-ranking model to leverage their complementary strengths.\footnote{As System-2 produces unordered results, we cannot directly apply an RRF re-ranker as in System-1.} Episodes, Entities (Categories), and Edges are re-ranked separately. These items are then reformatted into a unified memory context and provided to the answer model, together with the user query, to generate the final response.

\textbf{Practical Scalability.} To reduce the extra inference cost introduced by Global Selection, we employ \textbf{early stopping}, which terminates traversal once the model determines that all child nodes under the current branch are relevant, avoiding unnecessary deeper exploration.
\section{Experiments}
\subsection{Experiment Setups}
\textbf{Datasets.} We evaluate \M~on two well-known AI memory benchmarks: LoCoMo \citep{maharana2024evaluating} and LongMemEval-S \citep{wu2024longmemeval}.
LoCoMo consists of long-term conversations from 10 users, with each user contributing approximately 600 turns across 32 sessions, totaling around 16K tokens on average. The dataset contains roughly 2,000 questions spanning five diverse categories: Single-Hop, Multi-Hop, Temporal, Open-Domain, and Adversarial.
LongMemEval-S comprises 500 sessions, with each session containing one question and roughly 115K tokens, designed to evaluate five core memory abilities: information extraction, multi-session reasoning, temporal reasoning, knowledge updates, and abstention.

\begin{table*}[!th]
\centering
\small
\caption{Detailed performance (LLM-as-a-Judge score) on LoCoMo by question type. Following the common practice, Category 5 (Adversarial) is excluded from the results.}
\vspace{-.2cm}
\setlength\tabcolsep{0.1pt}
\begin{tabularx}{\textwidth}{Y l *{5}{Y}}
\toprule
\bf LLM & \bf Methods & \bf Multi-Hop & \bf Temporal & \bf Open-Domain & \bf Single-Hop & \bf Overall \\
\midrule
\bf $\sharp$Questions && 282 & 321 & 96 & 841 & 1540 \\
\midrule
\multirow{9}{*}{\rotatebox[origin=c]{90}{\textbf{GPT-4o-mini}}}
& Full Context & 66.8 & 56.2 & 48.6 & \underline{83.0} & 72.3 \\
& RAG & 59.9 & 62.9 & \underline{63.5} & 73.5 & 68.2 \\
\cmidrule{2-7}
& LangMem & 52.4 & 24.9 & 47.6 & 61.4 & 51.3 \\
& MemOS & 64.3 & 73.2 & 55.2 & 78.4 & 73.3 \\
& Mem0 & 60.3 & 50.4 & 40.6 & 68.1 & 61.3 \\
& Zep & 50.5 & 58.9 & 39.6 & 63.2 & 58.5 \\
& Nemori & 65.3 & 71.0 & 44.8 & 82.1 & 74.4 \\
& EMem-G & \underline{74.7} & \underline{76.0} & 57.3 & 82.3 & \underline{78.0} \\
& \bf \M & \bf 89.7 & \bf 77.6 & \bf 79.2 & \bf 95.7 & \bf 89.8 \\
\midrule
\multirow{11}{*}{\rotatebox[origin=c]{90}{\textbf{GPT-4.1-mini}}}
& Full Context & 77.2 & 74.2 & 56.6 & 86.9 & 80.6 \\
& RAG & 64.9 & 76.6 & 67.7 & 76.5 & 73.8 \\
\cmidrule{2-7}
& LangMem & 71.0 & 50.8 & 59.0 & 84.5 & 73.4 \\
& Mem0 & 68.2 & 56.9 & 47.9 & 71.4 & 66.3 \\
& Zep & 53.7 & 60.2 & 43.8 & 66.9 & 61.6 \\
& Nemori & 75.1 & 77.6 & 51.0 & 84.9 & 79.5 \\
& PREMem & 61.0 & 74.8 & 46.9 & 66.2 & 65.8 \\
& EverMemOS & 91.1 & 89.7 & 70.8 & 96.1 & 92.3 \\
& EMem-G & 79.6 & 80.8 & 71.7 & 90.5 & 85.3 \\
& \bf \M & \underline{91.8} & \underline{90.3} & \bf 82.3 & \underline{96.2} & \underline{93.3} \\
& \bf{\M}~($k$=30)& \bf 92.9 & \bf 90.7 & \underline{79.2} & \bf 97.1 & \bf 93.9 \\
\bottomrule
\end{tabularx}
\vspace{-.2cm}
\label{tab:locomo}
\end{table*}

\textbf{Baselines.} We compare \M~against the following baselines: LangMem \footnote{\url{https://github.com/langchain-ai/langmem}}, MemOS \citep{li2025memosmemoryosai}, Mem0 \citep{chhikara2025mem0buildingproductionreadyai}, Zep \citep{rasmussen2025zeptemporalknowledgegraph}, Nemori \citep{nan2025nemoriselforganizingagentmemory}, PreMem \citep{kim2025pre}, EverMemOS\footnote{\url{https://github.com/EverMind-AI/EverMemOS/}}, EMem-G \citep{zhou2025simplestrongbaselinelongterm} using GPT-4o-mini or GPT-4.1-mini as the backend model for memory building and question answering. We directly use their reported performance. In addition, we include two supplementary baselines: Full Context, which feeds the entire conversation history to the model, and RAG, which retrieves only episodes while keeping all other settings identical to \M. We also identified several other comparable baselines; however, due to missing details such as the backbone model and hyperparameter settings, we report their results only in \cref{sec:further_res}.

\textbf{Hyperparameters.} Following Nemori \citep{nan2025nemoriselforganizingagentmemory}, we limit the number of retrieved episodes in the answer prompt to top-$k=10$, while entities (including categories) and edges are limited to top-$2k=20$. We use Qwen3-Embedding-0.6B as the embedding model, with the embedding dimension fixed at 128 due to storage constraints. The re-ranker model used in the main experiments is Qwen3-Reranker-8B \citep{qwen3embedding}. We use neo4j\footnote{https://neo4j.com/} as the backend database. Across all experiments, the grader model is consistently GPT-4.1-mini to ensure accurate scoring.

\textbf{Metrics.} We employ LLM-as-a-Judge score (0/1) for evaluation and adopt the official judger prompt for each dataset. Following previous methods, Category 5 of LoCoMo is excluded from the final score.

\subsection{Experiment Results} \label{sec:exp_res}
The results can be found in \cref{tab:locomo,tab:lme}. Below, we provide detailed discussion on the results.

\textbf{Full-context models alone are insufficient for long-horizon AI memory.}
Across all settings, we observe a clear divergence between Full Context and RAG as context length grows. In LoCoMo, where the average context is roughly 16K tokens, which is well within the optimal operating window of modern LLMs (128K), the Full Context model remains competitive with most baselines. However, this behavior changes dramatically in LongMemEval-S, whose average context length reaches 115K tokens. As the input approaches or exceeds the model’s practical context limit, the Full Context model consistently degrades.
This contrast suggests an important implication for long-term memory: real deployments must support months or years of accumulated interaction history, far beyond what can be reliably handled by a single forward pass over the full context. Thus, relying solely on the model’s native context window without any additional memory management or retrieval mechanisms is insufficient for long-horizon, persistent AI memory systems.

\begin{table*}[!t]
\centering
\small
\caption{Detailed performance (LLM-as-a-Judge score) on LongMemEval-S, categorized by question type: single-session-user (SSU), multi-session (MS), single-session-preference (SSP), temporal reasoning (TR), knowledge update (KU), and single-session-assistant (SSA).}
\vspace{-.2cm}
\begin{tabularx}{\linewidth}{Y l *{7}{Y}}
\toprule
\bf LLM & \bf Methods & \bf SSU & \bf MS & \bf SSP & \bf TR & \bf KU & \bf SSA & \bf Overall \\
\midrule
\bf $\sharp$Questions && 70 & 133 & 30 & 133 & 78 & 56 & 500 \\
\midrule
\multirow{7}{*}{\rotatebox[origin=c]{90}{\textbf{GPT-4o-mini}}}
& Full Context & 78.6 & 38.3 & 6.7 & 42.1 & {78.2} & 89.3 & 55.0 \\
& RAG & 88.6 & 47.4 & \underline{70.0} & 63.2 & 70.5 & 91.1 & 67.2 \\
\cmidrule{2-9}
& Mem0 & 91.4 & {66.2} & 34.0 & 63.9 & 74.4 & \underline{96.4} & {71.1} \\
& Zep & \underline{92.9} & 47.4 & 53.3 & 54.1 & 74.4 & 75.0 & 63.2 \\
& Nemori & 88.6 & 51.1 & 46.7 & 61.7 & 61.5 & 83.9 & 64.2 \\
& EMem-G & 87.0 & \underline{73.6} & 32.2 & \underline{74.8} & \bf 94.4 & 87.5 & \underline{77.9} \\
& \bf \M & \bf 97.1 & \bf 76.7 & \bf 90.0 & \bf 83.5 & \underline{92.3} & \bf 100.0 & \bf 87.2 \\
\midrule
\multirow{8}{*}{\rotatebox[origin=c]{90}{\textbf{GPT-4.1-mini}}}
& Full Context & 85.7 & 51.1 & 16.7 & 60.2 & 76.9 & 98.2 & 65.6 \\
& RAG & 82.9 & 54.9 & 86.7 & 67.7 & 80.8 & 94.6 & 72.6 \\
\cmidrule{2-9}
& PREMem & 92.9 & 57.1 & 36.7 & 59.4 & 84.6 & 12.5 & 60.8 \\
& Mem0 & 94.3 & 66.9 & 86.7 & 75.9 & 87.2 & \underline{96.4} & 80.8 \\
& Nemori & 90.0 & 55.6 & 86.7 & 72.2 & 79.5 & 92.9 & 74.6 \\
& EverMemOS & \bf 100.0 & 78.5 & \underline{96.7} & 71.2 & 87.2 & 78.6 & 82.0 \\
& EMem-G & 94.8 & \underline{82.6} & 50.0 & \underline{83.7} & \bf 94.4 & 87.5 & \underline{84.9} \\
& \bf \M & \underline{98.6} & \bf 86.5 & \bf 100.0 & \bf 86.5 & \underline{93.6} & \bf 100.0 & \bf 91.6 \\
\bottomrule
\end{tabularx}
\label{tab:lme}
\vspace{-.2cm}
\end{table*}

\textbf{\M~consistently outperforms all baselines.} With limited and aligned context budget (10 episodes, 20 entities, and 20 edges), \M~achieves consistently superior performance across both benchmarks. For relatively easier tasks that are solvable within a single session or via single-hop reasoning, such as Single-Hop in LoCoMo and single-session-user, single-session assistant, and single-session-preference in LongMemEval-S, \M~reaches near-saturated scores. More importantly, on the challenging categories that require multi-hop evidence aggregation or complex temporal or event reasoning, \M~shows substantially larger margins over all baselines. These results demonstrate \M’s strong ability to organize and retrieve memory.

\begin{table*}[!t]
\centering
\small
\caption{Detailed LLM cost of \M~on LoCoMo using GPT-4.1-mini, reported as the \textbf{overall} number of prompt tokens, completion tokens, and end-to-end runtime for processing the entire dataset. Runtime depends heavily on database latency and parallelism configuration; the reported values are for reference only, and we will continue to optimize it for greater efficiency.}
\vspace{-.2cm}
\begin{tabularx}{\textwidth}{l*{4}{Y}}
\toprule
\bf Stage & \bf $\sharp$Prompt Tokens & \bf $\sharp$Completion Tokens & \bf E2E Runtime(s) \\
\midrule
Base Graph Ingestion & $3.87 \times 10^7$ & $1.06 \times 10^6$ & 1111.40 \\
Hierarchical Graph Ingestion & $1.39 \times 10^7$ & $9.27 \times 10^5$ & 3873.26 \\
\midrule
Global Selection & $1.37 \times 10^6$ & $1.21 \times 10^5$ & 3637.65 \\
\bottomrule
\end{tabularx}
\vspace{-.2cm}
\label{tab:cost}
\end{table*}

We also report the overall LLM cost of \M~when using GPT-4.1-mini to test LoCoMo in \cref{tab:cost}.
\subsection{Ablation Study}
To further assess the effectiveness of \M, we conduct comprehensive experiments from four perspectives: (1) the influence of System-1 and System-2 routes on the final results; (2) the effect of backend models (re-ranker, embedding model, LLM); (3) the impact of the top-$k$ parameter. For simplicity, these experiments are conducted on LoCoMo using GPT-4.1-mini.

\subsubsection{Influence of System-1 and System-2 Routing} \label{sec:route}
As stated in previous sections, System-1 Similarity Search provides a fast, heuristic retrieval mechanism based on similarities, while System-2 Global Selection performs a more structured and reflective selection process. To evaluate their individual and combined contributions, we compare three configurations: using only System-1, using only System-2, and using both jointly. For System-1, we further analyze four settings: (1) System-1 RAG: use retrieved episodes only; (2) System-1 Graph: use retrieved entities and edges only; (3) System-1 RAG + Graph: use episodes, entities and edges jointly; and (4) System-1 Re-ranked: the same to (3) but replace RRF re-ranker with Qwen3-Reranker-8B.

\begin{table*}[!th]
\centering
\small
\caption{Detailed performance (LLM-as-a-Judge score) on LoCoMo by question type.}
\vspace{-.2cm}
\begin{tabularx}{\textwidth}{l*{5}{Y}}
\toprule
\bf Settings & \bf Multi-Hop & \bf Temporal & \bf Open-Domain & \bf Single-Hop & \bf Overall \\
\midrule
\bf System-1 RAG & 64.9 & 76.6 & 67.7 & 76.5 & 73.8 \\
\bf System-1 Graph & 84.8 & 62.6 & 74.0 & 88.6 & 81.6 \\
\bf System-1 RAG + Graph & 85.1 & {84.7} & 75.0 & \underline{93.7} & \underline{89.1} \\
\bf System-1 Re-ranked & \underline{88.7} & \underline{85.0} & 75.0 & 92.4 & \underline{89.1} \\
\midrule
\bf System-2 Only & {88.1} & 78.5 & \underline{79.5} & 92.0 & 87.7 \\
\midrule
\bf System-1 + System-2 & \bf 91.8 & \bf 90.3 & \bf 82.3 & \bf 96.2 & \bf 93.3 \\
\bottomrule
\end{tabularx}
\vspace{-.2cm}
\label{tab:route}
\end{table*}

\begin{figure}[!th]
    \centering
    \includegraphics[width=\linewidth]{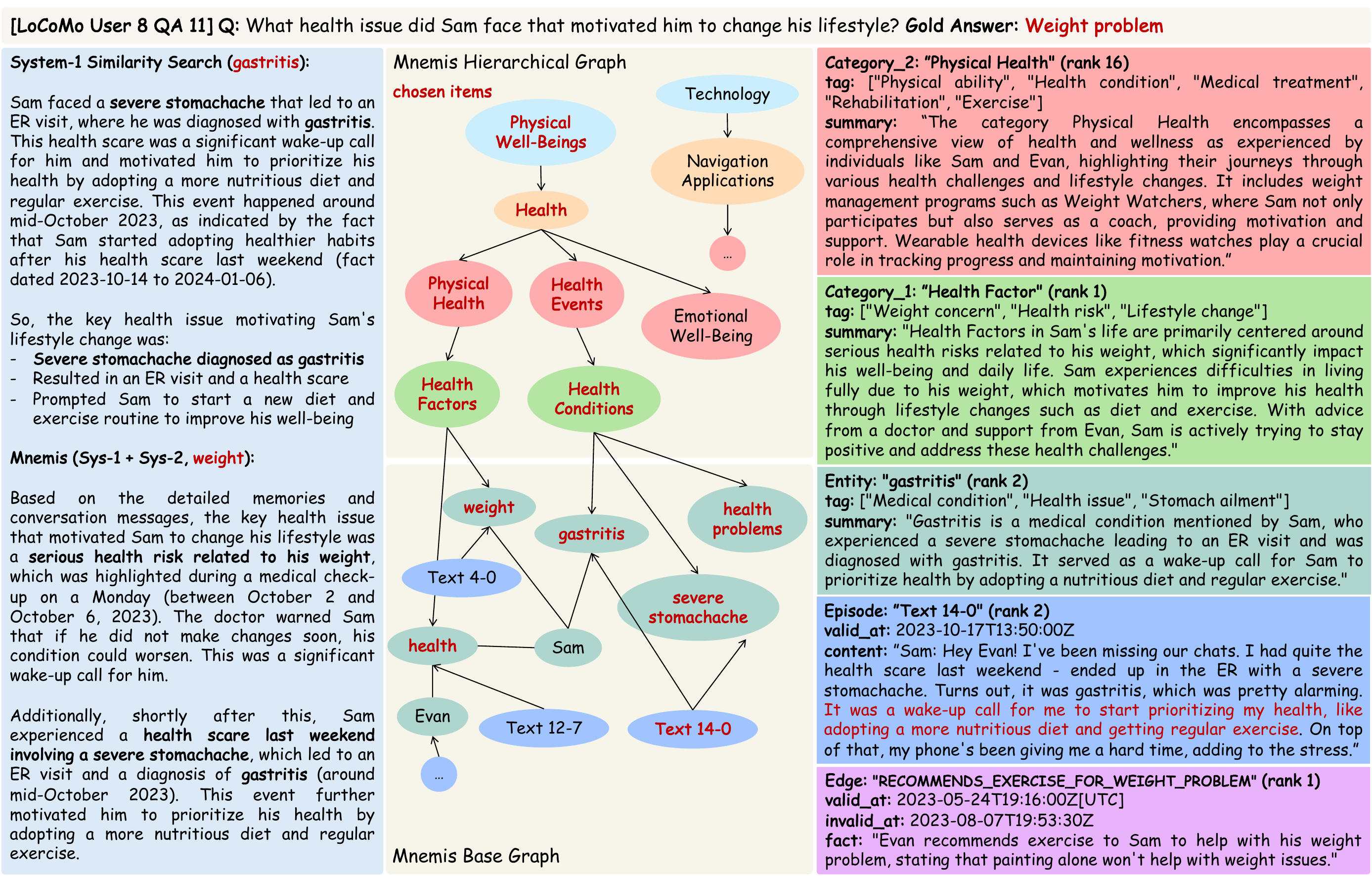}
    \vspace{-.2cm}
    \caption{\M~win case on LoCoMo benchmark. While similarity search fixates on the surface-level cause gastritis, \M successfully identifies the underlying root cause, namely that Sam is overweight. This condition leads to the gastritis and motivates him to change his lifestyle.}
    \label{fig:fig_case0}
    \vspace{-.2cm}
\end{figure}

The results can be found in \cref{tab:route}. System-1 Graph slightly wins System-1 RAG as the entities and edges are more condensed and informative compared to raw episodes. However, the compression comes at the cost of certain information loss, where episodes could compensate it. System-1 RAG + Graph hence achieves higher scores. The introduction of a re-ranking model affects performance in some sub-categories, but the overall score remains comparable to System-1 RAG + Graph. This indicates that the performance gain of System-1 + System-2 is not primarily driven by the re-ranking model, but instead stem from global selection.

According to the essence of the System-2 route, not all user queries are suitable for this process. We expect it to perform well on enumerative problems (\textit{e.g.}, "find all items that …") but weakly on temporal problems. Since the search query remains unchanged during the route, it may identify some key points but fail to capture the full sequence of temporal events. In the experiments, about 90.06\% (1387 / 1540) queries obtain the results, we hence report the average score on these valid queries. The results match our expectations. With both routes combined, all categories are improved, leveraging the complementary strengths of them.

To be more intuitive, we present \M~win cases from the LoCoMo and LongMemEval-S benchmarks to demonstrate the effectiveness of introducing System-2 Global Selection. As illustrated in \cref{fig:fig_case0}, when addressing the query "\textit{What health issue did Sam face that motivated him to change his lifestyle?}", similarity-based search could only retrieve "gastritis", merely a surface reason found in Episode Text 14-0. In contrast, equipped with Global Selection, we browse the hierarchical graph through "Physical Well-Beings" → "Health" → "Health Events" → "Health Conditions" → "gastritis" to locate the surface cause, and through "Physical Well-Beings" → "Health" → "Physical Health" → "Health Factors" → "weight" to locate the essential root cause. Along the search path, intermediate Categories, such as "Physical Health" and "Health Factors" naturally aggregate relevant information from their descendants in summary, which further enriches the retrieved context. For case from LongMemEval-S, please refer to \cref{sec:case}.

\subsubsection{Impact of top-$k$}

The top-$k$ parameter is introduced to balance answer cost and accuracy. In the main experiments, we use top-$k=10$, meaning that 10 episodes, 20 entities, and 20 edges are included in the context to generate the final answer. To evaluate the impact of top-$k$, we vary it across values of 5, 10, 30, and 50, and conduct experiments under the same configuration as \cref{sec:route}.

\vspace{-.2cm}
\begin{wrapfigure}{r}{0.5\textwidth}
    \centering
    \vspace{-.3cm}
    \includegraphics[width=\linewidth]{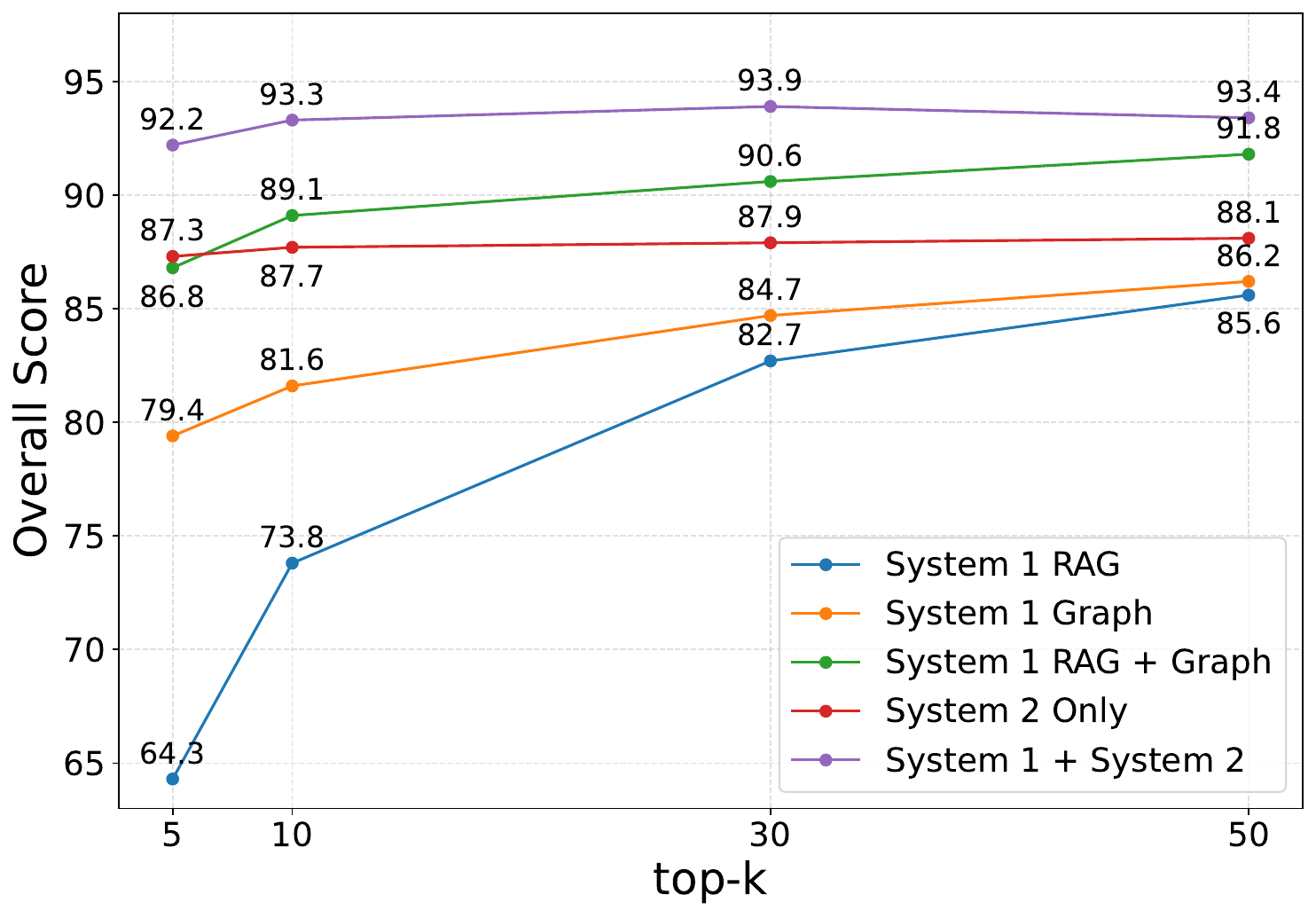}
    \vspace{-.7cm}
    \caption{LoCoMo result across different top-$k$ settings.}
    \vspace{-.3cm}
    \label{fig:fig_topk}
\end{wrapfigure}
\par\noindent

The results are shown in \cref{fig:fig_topk} with details in \cref{tab:topk}. Reducing top-$k$ from 10 to 5 leads to a clear performance drop in most settings, indicating that a top-$k$ of 5 is insufficient to capture all the evidence needed for user queries. System-1 RAG is especially sensitive to this reduction, particularly on Multi-Hop questions where diverse evidence across multiple parts of the history is required. In contrast, System-1 Graph, which retrieves more information-dense entities and edge, is less affected. System-1 RAG + Graph, which combines episodes, entities, and edges, shows more stable performance, and applying the re-ranker in System-2 Only or System-1 + System-2 further minimizes fluctuations.

\subsubsection{Effect of Backend Models}

In this section, we analyze the effect of backend models (LLM, re-ranker, embedding) on \M.

\textbf{LLM.} In \M, LLM is responsible for memory component extraction, de-duplication and retrieval. Switching LLM from GPT-4o-mini to GPT-4.1-mini, \M~shows clear improvements across all datasets and question types, as show in \cref{tab:locomo,tab:lme}. The same trend appears in the baseline methods, which suggests that the gains come from the stronger backend LLM rather than method specific factors. Given its favorable cost-performance trade-off, we recommend GPT-4.1-mini as the backend LLM for memory organization.

\textbf{Re-ranker.} Re-ranker organizes System-1 and System-2 search results to provide a compact context for the answer model. In the main experiments, we use Qwen3-Reranker-8B to obtain the best performance. We further evaluate \M~with two lightweight re-rankers: (1) Qwen3-Reranker-0.6B \citep{qwen3embedding} and (2) BGE-Reranker-V2-M3 (0.5B) \citep{chen2024bge}. As shown in \cref{tab:reranker}, replacing the re-ranker with these smaller models results in only minor performance regressions.

\textbf{Embedding Model.} The embedding model contributes to the similarity based search across \M. In our main experiments, we adopt Qwen3-Reranker-0.6B and reduce its embedding dimension from 1024 to 128 to control serving costs by using its MRL capability. To isolate and better understand the impact of embedding quality alone, we further evaluate System 1 RAG with three additional embedding models:
(1) BGE-M3 \citep{chen2024bge} with dimension 1024,
(2) all-MiniLM-L6-v2\footnote{https://huggingface.co/sentence-transformers/all-MiniLM-L6-v2} with dimension 384, and
(3) Gemma-300M \citep{embedding_gemma_2025} with dimension 768.
The results are presented in \cref{tab:embedder}.

\begin{table*}[!t]
\centering
\small
\caption{Detailed performance (LLM-as-a-Judge score) on LoCoMo by question type.}
\vspace{-.2cm}
\begin{tabularx}{\textwidth}{l*{5}{Y}}
\toprule
\bf Re-ranker & \bf Multi-Hop & \bf Temporal & \bf Open-Domain & \bf Single-Hop & \bf Overall \\
\midrule
Qwen3-Reranker-0.6B & 91.8 & 90.3 & 79.2 & 95.2 & 92.6 \\
BGE-Reranker-V2-M3 & 90.1 & 90.0 & 77.1 & 96.4 & 92.7 \\
Qwen3-Reranker-8B & 91.8 & 90.3 & 82.3 & 96.2 & 93.3 \\
\bottomrule
\end{tabularx}
\vspace{-.2cm}
\label{tab:reranker}
\end{table*}

\begin{table*}[!t]
\centering
\small
\caption{Detailed performance (LLM-as-a-Judge score) on LoCoMo by question type.}
\vspace{-.2cm}
\begin{tabularx}{\textwidth}{l*{5}{Y}}
\toprule
\bf Embedder & \bf Multi-Hop & \bf Temporal & \bf Open-Domain & \bf Single-Hop & \bf Overall \\
\midrule
MiniLM & 46.1 & 57.9 & 64.6 & 62.5 & 58.7 \\
BGE-M3 & 50.0 & 74.1 & 63.5 & 74.0 & 69.0 \\
Qwen3-Embedding-0.6B & 64.9 & 76.6 & 67.7 & 76.5 & 73.8 \\
Gemma-300M & 62.4 & 78.8 & 60.4 & 82.9 & 76.9 \\
\bottomrule
\end{tabularx}
\vspace{-.2cm}
\label{tab:embedder}
\end{table*}
\section{Related Work}
The core of AI memory lies in how they organize and retrieve past interactions. One straightforward way to organize LLMs’ memory is to treat them like individuals with hyperthymesia, supposing LLM can recall every past interaction without additional processing on historical messages. A line of work has therefore focused on enlarging the context window of LLMs \citep{liu2025comprehensivesurveylongcontext,peng2024yarn}. However, naively feeding the entire history can quickly become costly and inefficient in real-world applications due to the quadratic scaling of transformers with input length \citep{li2024retrieval,DBLP:conf/mm/Wu0C0LGKZ024,li2025tldrlongreweightingefficient,lin2025sigmadifferentialrescalingquery}, and irrelevant information in historical messages may further dilute the context \citep{shi2023large}.

Another line of work borrows the idea of episodic memory \citep{tulving1972episodic}. It stores historical messages as Episodes and only retrieves relevant items when dealing with user queries, termed as retrieval-augmented generation (RAG) \citep{arslan2024survey,lewis2020retrieval}. Graph-RAG, incorporating concepts from semantic memory \citep{tulving1972episodic}, extracts \textit{Entities} (key figures, objects, or concepts) and \textit{Edges} (events or relationships connecting them) and organizes memory into a structured graph  \citep{nan2025nemoriselforganizingagentmemory,chhikara2025mem0buildingproductionreadyai,wang2025mirixmultiagentmemoryllmbased,rasmussen2025zeptemporalknowledgegraph}. Some readers may note that GraphRAG \citep{edge2024local} introduces a hierarchy concept similar to \M. However, GraphRAG and \M~differ in two fundamental ways. First, GraphRAG constructs its hierarchy using community detection algorithms, where each lower-level node is assigned to a single parent. In contrast, \M~supports many-to-many mappings, allowing entities to belong to multiple higher-level categories and resulting in a more expressive and flexible hierarchy. Second, GraphRAG generates answers by independently querying each community and aggregating the outputs into a final response. In \M, the model instead performs top-down hierarchical browsing to retrieve relevant memory components, and the final answer is produced based on the aggregated memory context rather than separate community-specific responses.
\section{Conclusion}
In this work, we introduce \M, a unified memory framework to organize and retrieve AI memory. By combining a refined base graph for System-1 Similarity Search with a hierarchical graph designed to support System-2 Global Selection, \M~enables more accurate retrieval than existing RAG and Graph-RAG approaches on memory benchmarks, achieving 93.9 in LoCoMo and 91.6 on LongMemEval-S. While the results are strong, several important directions remain open. In future work, we plan to support more data modalities and  enhance global selection with more flexible graph traversal and planning mechanisms.
\bibliographystyle{assets/plainnat}
\bibliography{reference}
\clearpage
\appendix
\section{Case Study} \label{sec:case}
\begin{figure}[!th]
    \centering
    \includegraphics[width=\linewidth]{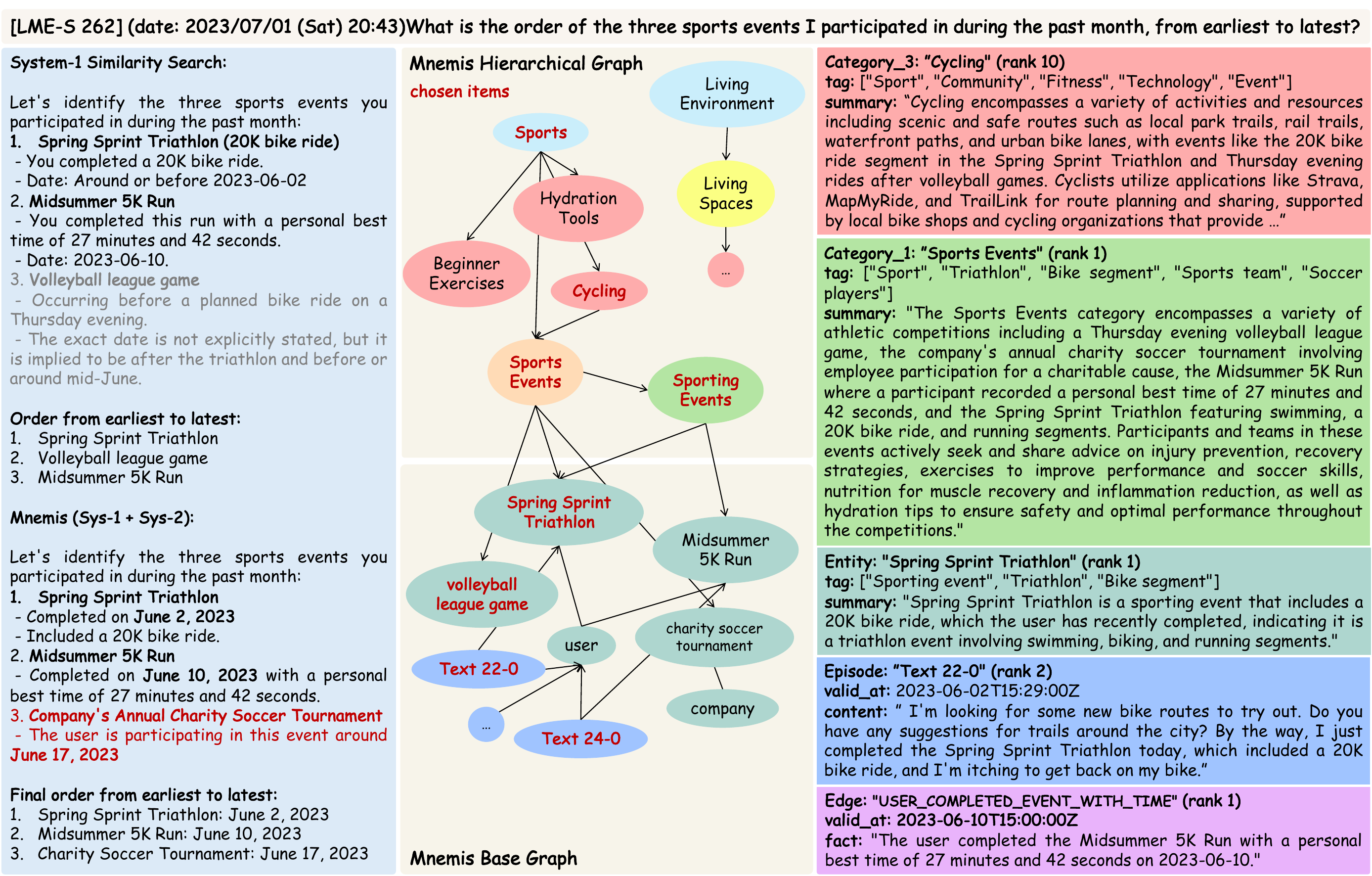}
    \caption{\M~win case on LongMemEval-S benchmark, where \M~successfully retrieve "Company's Annual Charity Soccer Tournament" from "Sports Events".}
    \label{fig:fig_case1}
    \vspace{-3pt}
\end{figure}

\cref{fig:fig_case1} provides another win case of \M~on LongMemEval-S. This question asks the LLM to order the three sports events the user participated in during June 2026. Similarity-based retrieval struggles here, as a limited top-$k$ often fails to surface all relevant sports events. In contrast, \M~can simply start from top-leevel category "Sports" and browse down through Category "Sports Events" and "Sporting Events" and then reliably retrieve all events.

Although the ground truth entities "Midsummer 5K Run" and "charity soccer tournament" are filtered out in the final answer context due to limited top-$k$, their related edges, such as "The user participates in the company's annual charity soccer tournament. (2023-06-17 - now)" and "The user completed a 5K run at the Midsummer 5K Run with a personal best time of 27 minutes and 42 seconds. (2023-06-10 - now)" are still retrieved with rank 9 and 2 respectively. Besides, category nodes like "Sports Events", summarizing the content of their children nodes, also provide sufficient content for the model to answer the question correctly.

\section{Further Benchmark Results} \label{sec:further_res}

\subsection{Evaluation Metrics}
In the main experiments, we adopt LLM-as-a-Judge (LAJ) as the primary evaluation metric for comparing \M~against all baselines, following common practice.
We directly use the official judging prompt to ensure comparability.
To further assess the reliability of LAJ, we conduct a small-scale human evaluation.

\begin{table}[ht]
\centering
\caption{Sampled Question ID for Human Evaluation in LoCoMo.}
\label{tab:sampled-qa}
\small
\begin{tabular}{lcccccccccc}
\toprule
\bf User IDs & 0 & 1 & 2 & 3 & 4 \\
\midrule
\bf LLM-as-a-Judge True Question IDs & 31, 8, 74 & 13, 74, 11 & 7, 26, 60 & 6, 160, 59 & 124, 161, 80 \\
\bf LLM-as-a-Judge False Question IDs & 7, 138, 137 & 73, 47, 20 & 17, 49, 63 & 195, 92, 53 & 165, 166, 6 \\
\midrule
\bf User IDs & 5 & 6 & 7 & 8 & 9 \\
\midrule
\bf LLM-as-a-Judge True Question IDs & 104, 110, 23 & 47, 65, 96 & 25, 97, 94 & 126, 147, 32 & 99, 156, 54 \\
\bf LLM-as-a-Judge False Question IDs & 55, 19, 119 & 1, 62, 31 & 117, 77, 23 & 102, 39, 70 & 7, 1, 61 \\
\bottomrule
\end{tabular}
\end{table}

Specifically, we manually sample \M~'s outputs on LoCoMo, inspecting both win and loss cases (three labeled correct and three labeled incorrect per user), resulting in 60 questions sampled from 1,540 evaluation instances, as shown in \cref{tab:sampled-qa}. Human evaluation is performed using the same grading criteria as LoCoMo. Across these samples, we did not observe any misalignment between the LLM judge and human evaluation. We include the detailed judging results in our GitHub repository.

We also note several strong recently proposed baselines: MIRIX \citep{wang2025mirixmultiagentmemoryllmbased}, MemU\footnote{\url{https://github.com/NevaMind-AI/memU}}, EmergenceMem\footnote{\url{https://www.emergence.ai/blog/sota-on-longmemeval-with-rag}}. However, due to incomplete implementation details, such as backend model configuration or system workflow, we could not obtain their results under comparable settings. To maintain fairness, we therefore exclude them from the main evaluation. Here, we report their best publicly available performance, regardless of configuration differences, to provide readers with a broader view of the current landscape.

The full results are reported in \cref{tab:locomo_full,tab:lme_full} and align with the findings summarized in \cref{sec:exp_res}. \M~consistently outperforms all baseline methods across both benchmarks.

\begin{table*}[!t]
\centering
\small
\caption{Detailed performance (LLM-as-a-Judge score) on LoCoMo by question type. Following the common practice, Category 5 (Adversarial) is excluded from the results.}
\vspace{1pt}
\begin{tabularx}{\textwidth}{l *{5}{Y}}
\toprule
\bf Methods & \bf Multi-Hop & \bf Temporal & \bf Open-Domain & \bf Single-Hop & \bf Overall \\
\midrule
\bf $\sharp$Questions & 282 & 321 & 96 & 841 & 1540 \\
\midrule
Full Context & 77.2 & 74.2 & 56.6 & 86.9 & 80.6 \\
RAG & 64.9 & 76.6 & 67.7 & 76.5 & 73.8 \\
\midrule
LangMem & 71.0 & 50.8 & 59.0 & 84.5 & 73.4 \\
Mem0 & 68.2 & 56.9 & 47.9 & 71.4 & 66.3 \\
Zep & 53.7 & 60.2 & 43.8 & 66.9 & 61.6 \\
Nemori & 75.1 & 77.6 & 51.0 & 84.9 & 79.5 \\
PREMem & 61.0 & 74.8 & 46.9 & 66.2 & 65.8 \\
EMem-G & 79.6 & 80.8 & 71.7 & 90.5 & 85.3 \\
MIRIX & 83.7 & 88.4 & 65.6 & 85.1 & 85.4 \\
EverMemOS & 91.1 & 89.7 & 70.8 & 96.1 & 92.3 \\
MemU & 88.3 & \bf 92.5 & 77.1 & 94.9 & 92.1 \\
\bf{\M}& \bf 92.9 & 90.7 & \bf{79.2} & \bf 97.1 & \bf 93.9 \\
\bottomrule
\end{tabularx}
\label{tab:locomo_full}
\end{table*}

\begin{table*}[!t]
\centering
\small
\caption{Detailed performance (LLM-as-a-Judge score) on LongMemEval-S, categorized by question type: single-session-user (SSU), multi-session (MS), single-session-preference (SSP), temporal reasoning (TR), knowledge update (KU), and single-session-assistant (SSA).}
\vspace{1pt}
\begin{tabularx}{\linewidth}{l *{7}{Y}}
\toprule
\bf Methods & \bf SSU & \bf MS & \bf SSP & \bf TR & \bf KU & \bf SSA & \bf Overall \\
\midrule
\bf $\sharp$Questions & 70 & 133 & 30 & 133 & 78 & 56 & 500 \\
\midrule
Full Context & 85.7 & 51.1 & 16.7 & 60.2 & 76.9 & 98.2 & 65.6 \\
RAG & 82.9 & 54.9 & 86.7 & 67.7 & 80.8 & 94.6 & 72.6 \\
\midrule
PREMem & 92.9 & 57.1 & 36.7 & 59.4 & 84.6 & 12.5 & 60.8 \\
Mem0 & 94.3 & 66.9 & 86.7 & 75.9 & 87.2 & {96.4} & 80.8 \\
Nemori & 90.0 & 55.6 & 86.7 & 72.2 & 79.5 & 92.9 & 74.6 \\
EverMemOS & \bf 100.0 & 78.5 & {96.7} & 71.2 & 87.2 & 78.6 & 82.0 \\
EMem-G & 94.8 & {82.6} & 50.0 & {83.7} & \bf 94.4 & 87.5 & {84.9} \\
EmergenceMem & 98.6 & 81.2 & 60.0 & 85.7 & 83.3 & \bf 100.0 & 86.0 \\
\bf \M & {98.6} & \bf 86.5 & \bf 100.0 & \bf 86.5 & {93.6} & \bf 100.0 & \bf 91.6 \\
\bottomrule
\end{tabularx}
\label{tab:lme_full}
\end{table*}

\section{Detailed Performance Impact of top-$k$}

Beyond the overall trend as shown in \cref{fig:fig_topk}, a closer breakdown by question type in \cref{tab:topk} reveals distinct behavioral patterns across retrieval strategies. For System-1 RAG, increasing top-$k$ consistently improves performance across all categories, with particularly large gains on Multi-Hop subset (49.6→81.6). This suggests that the retrieved text contains necessary but scattered evidence, and restricting retrieval too aggressively leads to missing critical evidence. However, even at high top-$k$, RAG remains relatively weak on Multi-Hop (peaking at 81.6), indicating difficulty in identifying temporally aligned evidence without explicit structure.

In contrast, System-1 Graph shows a stronger starting point, especially on structured attributes (e.g., Single-Hop: 86.7 with top-$k=5$), meaning the graph format inherently surfaces salient relational information without requiring large retrieval volumes. Yet, the improvement curve is flatter compared to RAG, especially for Temporal and Open-Domain questions, where explicit structural relations help but cannot fully compensate for missing richer semantic context.

When combining both storage types in System-1 RAG + Graph, the benefits become additive: performance improves steadily and remains stable even under lower top-$k$ settings. Notably, Multi-Hop and Temporal queries benefit the most from this hybrid storage design (e.g., 83.2 vs. 61.4/70.1 at top-$k=5$), demonstrating that structured and unstructured information provide complementary retrieval signals.

Finally, applying System-2 and re-ranker, either alone or on top of System-1, further suppresses top-$k$ sensitivity. System-1 + System-2 consistently delivers the highest and most stable performance (92.2–93.9 Overall), showing that even when overly large or overly sparse retrieval occurs, the re-ranker filters noise and prioritizes the most relevant evidence.
\begin{table*}[!th]
\centering
\small
\caption{Detailed performance (LLM-as-a-Judge score) on LoCoMo by question type.}
\vspace{1pt}
\setlength\tabcolsep{.1pt}
\begin{tabularx}{\textwidth}{l*{6}{Y}}
\toprule
\bf Settings & \bf top-$k$ &\bf Multi-Hop & \bf Temporal & \bf Open-Domain & \bf Single-Hop & \bf Overall \\
\midrule
\multirow{4}{*}{\bf System-1 RAG}
& 5 & 49.6 & 70.1 & 65.6 & 66.8 & 64.3 \\
& 10 & 64.9 & 76.6 & 67.7 & 76.5 & 73.8 \\
& 30 & 77.3 & 82.9 & 69.8 & 86.0 & 82.7 \\
& 50 & 81.6 & 84.1 & 71.9 & 89.1 & 85.6 \\
\midrule
\multirow{4}{*}{\bf System-1 Graph}
& 5 & 79.4 & 61.4 & 75.0 & 86.7 & 79.4 \\
& 10 & 84.8 & 62.6 & 74.0 & 88.6 & 81.6 \\
& 30 & 87.6 & 66.4 & 77.1 & 91.6 & 84.7 \\
& 50 & 90.1 & 68.5 & 79.2 & 92.4 & 86.2 \\
\midrule
\multirow{4}{*}{\bf System-1 RAG + Graph}
& 5 & 81.6 & 83.2 & 71.9 & 91.6 & 86.8 \\
& 10 & 85.1 & 84.7 & 75.0 & 93.7 & 89.1 \\
& 30 & 87.9 & 86.3 & 76.0 & 94.9 & 90.6 \\
& 50 & 89.4 & 86.9 & 78.1 & 96.0 & 91.8 \\
\midrule
\multirow{4}{*}{\bf System-2 Only}
& 5 & 84.4 & 81.9 & 79.5 & 91.1 & 87.3 \\
& 10 & 88.1 & 78.5 & 79.5 & 92.0 & 87.7 \\
& 30 & 86.1 & 81.3 & 78.3 & 98.3 & 87.9 \\
& 50 & 88.1 & 80.2 & 77.1 & 92.2 & 88.1 \\
\midrule
\multirow{4}{*}{\bf System-1 + System-2}
& 5 & 88.3 & 89.1 & 82.3 & 95.8 & 92.2 \\
& 10 & 91.8 & 90.3 & 82.3 & 96.2 & 93.3 \\
& 30 & 92.9 & 90.7 & 79.2 & 97.1 & 93.9 \\
& 50 & 92.2 & 90.3 & 81.3 & 96.3 & 93.4 \\
\bottomrule
\end{tabularx}
\label{tab:topk}
\end{table*}


\section{Prompts} \label{sec:prompt}

\lstset{
  language=Python,                  
  numbers=left,                     
  numberstyle=\small\color{gray},   
  stepnumber=1,                     
  breaklines=true,                  
  breakatwhitespace=false,          
  frame=single,                     
  tabsize=2,                        
  basicstyle=\ttfamily\small,       
  keywordstyle=\color{blue},        
  stringstyle=\color{orange},       
  commentstyle=\color{green!50!black}, 
  moredelim=**[is][\bfseries\color{red}]{@}{@}
}

To implement our results, we release the key prompts in our procedure. Below is the instruction to build hierarchical graph.
\begin{lstlisting}
def extract_category_nodes(context: dict[str, Any], layer: int, prev_example: str) -> list[Message]:
    sys_prompt = f"""You are an AI assistant specialized in semantic categorization of nodes.
# INSTRUCTIONS:

You are given a list of node names, each prefixed with an index, each followed with a brief description of the name (e.g., 1. dog: [domestic animal]).
Your task is to:

1. Group the nodes into semantically meaningful categories based on shared attributes, considering both inherent characteristics of the node names and the DESCRIPTIONS of the nodes, NOT relying solely on the DESCRIPTIONS.
   All EXISTING CATEGORIES are provided for you.

   - If a node's attribute matches an existing category, it should be added under that category.
   - If a node name has attributes that do not match any existing category, create a new category and add it.
   - The category name MUST NOT include the word "and" as a connector.

     Examples of INVALID categories:
         - "Food and Drinks"
         - "University and Courses"

     Examples of VALID categories:
         - "Food"
         - "Drinks"
         - "University"
         - "Courses"

2. Output each category as a dictionary entry where the key is the category name and the value is a list of node indexes (integers). Only refer to nodes by their indexes. Do not repeat node names.

   Output format:
   [
     {{"category": "xx", "indexes": [0, 1, 2, 4]}},
     {{"category": "xxx", "indexes": [2, 3, 4]}}
   ]

   The tag is a list of descriptors (each descriptor maximum 3 words, maximum 5 descriptors) that concisely captures the nature or type of the node.

   Tag example:
     - Entity name: "Son"
     - Tag: ["Family member", "Happy kid", "Anime lover"]

3. A node CAN be assigned to MULTIPLE categories at the same time.

   Key points for multi-category classification:
   - Each item can be assigned to multiple categories based on shared attributes.
   - When multiple categories are formed for an item, select the minimal subset of features that are common across the grouped items.

   Examples for different hierarchy levels:

   Layer 1 (specific entities):
   - "Microsoft Research Asia" and "Microsoft Research Shanghai" share the same parent organization (Microsoft) and a similar research focus (AI). They are grouped under:
       - "Microsoft Research Labs"
       - "AI-focused Research Labs"

   - "Microsoft Research Asia" belongs to both "Microsoft Research Labs" and "NLP-focused Labs".

   Layer 2 (category nodes from Layer 1):
   - "Microsoft Research Labs" belongs to:
       - "Tech Company Labs"
       - "AI Research Organizations"

   - "University AI Labs" belongs to:
       - "Academic Institutions"
       - "AI Research Organizations"

   Layer 3 (higher-level abstractions):
   - "Tech Company Labs" belongs to:
       - "Commercial Organizations"
       - "Research Institutions"

   - "Academic Institutions" belongs to:
       - "Educational Organizations"
       - "Research Institutions"

   Layer 4 (top-level concepts):
   - "Research Institutions" belongs to:
       - "Knowledge Organizations"

   - "Commercial Organizations" belongs to:
       - "Economic Entities"

4. There must be NO leftover or ungrouped nodes. Single-member categories are allowed if necessary.

5. The node name "user" and any first-person references ("I", "me") MUST be categorized into one category called "Speaker".
"""

    guidance = f"""
<GUIDANCE ON CATEGORY GRANULARITY>
You are performing hierarchical semantic clustering from specific to abstract.

You are currently at Layer {layer}, where:
- Layer 1 contains the most specific, fine-grained categories.
- Higher layers should group lower-layer categories into broader, more abstract super-categories.

Example:

Layer 1:
- "Golden Retriever", "Poodle", "German Shepherd" -> "Dog breeds"
- "Persian Cat", "Siamese Cat" -> "Cat breeds"
- "Bengal Tiger", "Siberian Tiger" -> "Tiger subspecies"
- "Oak tree", "Pine tree" -> "Tree species"

Layer 2:
- "Dog breeds", "Cat breeds" -> "Pets"
- "Dog breeds", "Tiger subspecies" -> "Mammals"
- "Tiger subspecies" -> "Wild animals"
- "Tree species" -> "Trees"

Layer 3:
- "Pets", "Wild animals" -> "Animals"
- "Trees" -> "Plants"

Layer 4:
- "Animals", "Plants" -> "Living organisms"

Key points:
- Categories may belong to multiple parent categories.
- Do not merge categories that are too loosely related.

Your job at Layer {layer}:
- Merge semantically similar categories from Layer {layer - 1}.
- Each new category should reflect a shared attribute, domain, or higher-level concept.
- Multiple category assignments are allowed when justified.

Previous Layer {layer - 1} categories example:
{prev_example}
</GUIDANCE ON CATEGORY GRANULARITY>
"""

    user_prompt = f"""
<NODE INDEXED NAMES AND DESCRIPTIONS>
{context['content']}
</NODE INDEXED NAMES AND DESCRIPTIONS>

<EXISTING CATEGORIES>
These are names and descriptions of categories previously created. Reuse them if applicable.
{context['existing_categories']}
</EXISTING CATEGORIES>

{guidance}

# ATTENTION
- The node name "user" and any first-person references ("I", "me") MUST be categorized into one category called "Speaker". If the "Speaker" category does not exist, skip this node.
- The category name MUST NOT include the word "and".

Please follow the INSTRUCTIONS and GUIDANCE carefully to ensure accurate categorization and meaningful hierarchical relationships.
DO NOT INCLUDE ANY INVALID CATEGORIES.
"""

    return [
        Message(role="system", content=sys_prompt),
        Message(role="user", content=user_prompt),
    ]
\end{lstlisting}

Below is the instruction to conduct Global Selection.
\begin{lstlisting}
NODE_SELECTION_PROMPT_TEMPLATE = """You are analyzing a hierarchical knowledge graph to help answer a user query.

Select all nodes that could help answer the query. A node is helpful if it:

- Directly relates to the query;
- Covers a clearly relevant topic, concept, or category;
- Provides useful background or context;
- Contains user-specific information (e.g. interests, goals, constraints);
- Likely has sub-nodes that may be helpful.

Do not be overly strict: include nodes that might provide context or personalization, even if they seem partially redundant.

For each selected node:
- "name" is the node's name.
- "uuid" is the node's unique identifier.
- "get_all_children" is an boolean value. Set true only if you're confident all its sub-nodes are helpful.
---
User Query:
"{query}"

Available Nodes:
{nodes_info}
"""
\end{lstlisting}
\clearpage
\end{document}